\PassOptionsToPackage{table}{xcolor}
\PassOptionsToPackage{pagebackref,breaklinks=true,colorlinks,citecolor=blue,urlcolor=blue,linkcolor=blue,bookmarks=false}{hyperref}
\PassOptionsToPackage{noabbrev,nameinlink,capitalize}{cleveref}

\documentclass[onecolumn, numbers]{april_aigc}

\definecolor{codekw}{rgb}{0.13,0.13,1}

\usepackage[utf8]{inputenc} 
\usepackage{url}            
\usepackage{amsfonts}       
\usepackage{amssymb}        
\usepackage{amsmath}        
\usepackage{nicefrac}       
\usepackage{microtype}      
\usepackage{xspace}         
\usepackage{fix-cm}
\usepackage[T1]{fontenc}

\usepackage{booktabs}       
\usepackage{wrapfig}        
\usepackage{multicol}       
\usepackage{multirow}       
\usepackage{makecell}       
\usepackage{tabularx}       
\usepackage{adjustbox}      
\usepackage{colortbl}       
\usepackage{pifont}         
\usepackage{enumitem}
\usepackage[normalem]{ulem}
\usepackage{listings}
\usepackage[most]{tcolorbox}
\usepackage{xcolor}

\definecolor{yamlbg}{RGB}{248,248,248}
\definecolor{yamlkey}{RGB}{33,74,135}
\definecolor{yamlcomment}{RGB}{0,120,0}
\usepackage{xcolor}
\definecolor{linkcolor}{named}{aprilblue}
\definecolor{urlcolor}{RGB}{255,105,180}
\definecolor{citecolor}{RGB}{66,168,235}
\definecolor{lightgray}{rgb}{0.8, 0.8, 0.8}
\definecolor{darkgreen}{rgb}{0.00, 0.81, 0.78}

\definecolor{gray_tab}{RGB}{220, 220, 220}
\definecolor{blue_tab}{RGB}{227, 240, 251}
\definecolor{oran_tab}{RGB}{252, 242, 237}
\definecolor{whit_tab}{RGB}{255, 255, 255}
\definecolor{green_code}{RGB}{55, 126, 34}

\usepackage{algorithm}
\usepackage{algorithmic}
\usepackage{listings}
\usepackage{etoolbox}

\makeatletter
\AfterEndEnvironment{algorithm}{\let\@algcomment\relax}
\AtEndEnvironment{algorithm}{\kern2pt\hrule\relax\vskip3pt\@algcomment}
\let\@algcomment\relax
\newcommand\algcomment[1]{\def\@algcomment{\footnotesize#1}}
\renewcommand\fs@ruled{\def\@fs@cfont{\bfseries}\let\@fs@capt\floatc@ruled
  \def\@fs@pre{\hrule height.8pt depth0pt \kern2pt}%
  \def\@fs@post{}%
  \def\@fs@mid{\kern2pt\hrule\kern2pt}%
  \let\@fs@iftopcapt\iftrue}
\makeatother

\newcommand{\cmark}{\ding{52}\xspace}%
\newcommand{\xmark}{\ding{56}\xspace}%

\def\onedot{.\xspace}

\def\etc{\textit{etc}\onedot}

\usepackage{natbib}
\usepackage[pagebackref,breaklinks=true,colorlinks,citecolor=blue,urlcolor=blue,linkcolor=blue,bookmarks=false]{hyperref}
\AtEndPreamble{
    \usepackage[capitalize]{cleveref}
    \crefname{section}{Sec.}{Secs.}
    \Crefname{section}{Section}{Sections}
    \crefname{table}{Tab.}{Tabs.}
    \Crefname{table}{Table}{Tables}
    \crefname{equation}{Eq.}{Eqs.}
    \Crefname{equation}{Equation}{Equations}
    \crefname{figure}{Fig.}{Figs.}
    \Crefname{figure}{Figure}{Figures}
}
\hypersetup{colorlinks=true,linkcolor=linkcolor,urlcolor=urlcolor,citecolor=citecolor}

\usepackage{caption}
\DeclareCaptionFormat{custom}{{\color{aprilblue}\sffamily\textbf{#1 #2}} #3}
\titleformat*{\section}{\color{aprilblue}\Large\sffamily\bfseries}
\titleformat*{\subsection}{\color{aprilblue}\large\sffamily\bfseries}
\titleformat*{\subsubsection}{\color{aprilblue}\normalsize\sffamily\bfseries}

\usepackage{fancyhdr}
\newif\ifshowlogo
\showlogotrue   
\newcommand{\insertlogo}{%
  \ifshowlogo
    \IfFileExists{assets/april_logo1.png}%
    {\includegraphics[height=0.68cm]{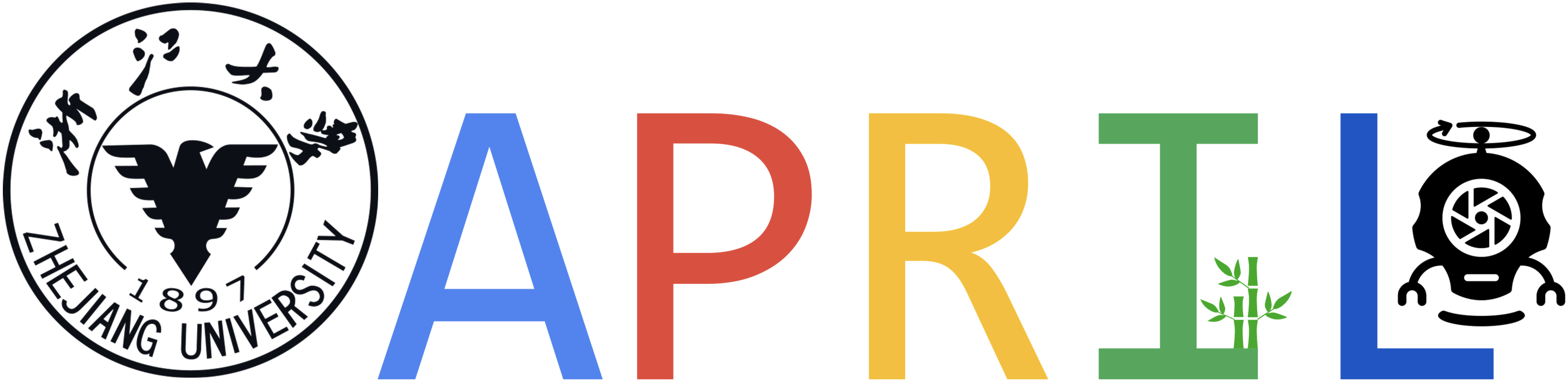}}
    {}%
  \fi
}
\newif\ifshowtoc
\showtocfalse

\renewcommand{\title}[1]{\def\titlelist{{\fontsize{20pt}{28pt}\selectfont\sffamily\bfseries #1}}}
\title{APRIL-MedSeg: A Modular Medical Image \ Segmentation Toolbox Embracing Modern Paradigms}

\author[1]{Juntao Jiang}
\author[1]{Jinsheng Bai}
\author[2]{Linxuan Fan}
\author[3]{Yali Bi}
\author[1\dagger]{Jiangning Zhang}
\author[1,\dagger]{Yong Liu}

\affiliation[1]{Zhejiang University, APRIL Lab}
\affiliation[2]{Vanderbilt University}
\affiliation[3]{University of Science and Technology of China}

\contribution[\dagger]{Corresponding author}

\abstract{
We present \textbf{APRIL-MedSeg}, a YAML-driven modular framework for 2D medical image segmentation. It provides a unified and extensible ecosystem that decomposes segmentation networks into reusable components. Also, the framework integrates a broad spectrum of advanced paradigms, including semi-supervised learning, domain adaptation, knowledge distillation, weakly supervised learning, and text-guided segmentation as well as foundation model support. A registry-based configuration system with inheritance enables flexible and reproducible experiment management, supporting seamless switching across models, datasets, and training strategies. In addition, the framework provides a unified interface for medical datasets, augmentation pipelines, deployment utilities and model ensembling. Overall, APRIL-MedSeg is designed as a general-purpose research and development platform that bridges algorithmic innovation and practical deployment, while also serving as a structured ecosystem for systematically organizing and reproducing advances in medical image segmentation. The code is available under an Apache 2.0 license.
}

\coverdate{\today}
\covercorrespondence{\email{yongliu@iipc.zju.edu.cn},\email{186368@zju.edu.cn}}
\coversourcecode{https://github.com/juntaoJianggavin/APRIL-MedSeg}

\begin{document}

\maketitle
\thispagestyle{plain}
\section{Introduction}

Medical image segmentation, the precise, pixel-level delineation of anatomical structures and pathological regions, stands as the fundamental cornerstone of modern medical image analysis. By transforming complex, unstructured imaging data into actionable quantitative insights, it establishes an indispensable structural foundation for a multitude of downstream clinical workflows. Crucially, high-fidelity segmentation is not merely an analytical step; it is the critical prerequisite for early disease diagnosis, personalized treatment planning, image-guided interventions, and longitudinal disease monitoring. Ultimately, it bridges the vital gap between raw diagnostic imaging and precision medicine, directly empowering clinical decision-making and improving patient outcomes.

Historically, general-purpose vision toolboxes such as MMSegmentation~\cite{mmseg2020}, PaddleSeg~\cite{liu2021paddleseg}, and the 2D-focused Segmentation Models PyTorch (SMP)~\cite{Iakubovskii:2019} played a monumental role in democratizing segmentation techniques through modular design and rich model zoos. However, as pure computer vision paradigms shift, the architectural evolution of these foundational frameworks has largely plateaued. In response to the unique complexities of clinical data, the community has pivoted towards domain-specific platforms like MONAI~\cite{cardoso2022monai}, MIST~\cite{celaya2024mist}, and PaddleSeg-MedicalSeg. While these medical-centric ecosystems remain highly active, excelling in clinical preprocessing, 3D volumetric standardization, and end-to-end evaluation, they predominantly focus on establishing reliable conventional pipelines. Consequently, a critical gap persists: neither legacy general-vision toolboxes nor contemporary medical frameworks can seamlessly accommodate three major frontier trends. Specifically, they struggle to unify (1) next-generation efficient architectures (e.g., State Space Models), (2) modern training paradigms, and (3) the deep integration of foundation models. Currently, no existing framework provides this unification within a configuration-driven, highly reproducible system.

While the recent proliferation of AI-assisted coding tools has led some to question the continued necessity of dedicated segmentation frameworks, this view fundamentally misjudges the rigor required in medical AI research. Writing isolated scripts is vastly different from conducting systematic clinical validation. A robust, actively maintained framework serves three irreplaceable functions: (1) it abstracts away low-level engineering complexities, decoupling system design from algorithmic innovation so researchers can rapidly prototype next-generation architectures; (2) it enforces standardized, highly reproducible evaluation under strict clinical data and training protocols; and (3) it cultivates a unified ecosystem that consolidates fragmented methodologies, preventing redundant re-implementation and accelerating community-driven progress.

For an entire generation of researchers, seminal toolboxes like MMSegmentation and PaddleSeg were not merely software; they were the practical textbooks through which we first navigated the intricacies of deep learning. We hold a profound, enduring respect for these pioneering ecosystems that shaped our foundational understanding. Yet, the inexorable march of the field demands infrastructure that can adapt without friction. Standing on the shoulders of these beloved predecessors, APRIL-MedSeg addresses contemporary gaps with the following contributions:

\begin{itemize}[nosep,leftmargin=*]
    \item A \textbf{four-module free-combination design} decomposing segmentation networks into encoder, decoder, skip connection, and bottleneck. Each is independently interchangeable via 6 registries, enabling $177 \times 45 \times 25 \times 17$ potential combinations;
    \item \textbf{130 completed architectures} covering 8 paradigm families with state-of-the-art methods up to 2026, plus 177 encoders (including 39 vision or multi-modal foundation models across 9 modalities);
    \item \textbf{Five advanced paradigms} totaling 97 methods that address annotation scarcity, domain shift, model compression, weak label challenges, and text-guided methods in multi-modal settings;
    \item \textbf{917 YAML configurations} with inheritance-based composition, enabling zero-code switching between any model, paradigm, or dataset combination;
    \item \textbf{Comprehensive data and deployment infrastructure}: featuring 25 datasets, 24 augmentations, ONNX export, FLOPs/FPS profiling, Test-Time Augmentation (TTA), and ensemble strategies.
\end{itemize}

\begin{figure*}[t]
    \centering
    \includegraphics[width=\textwidth]{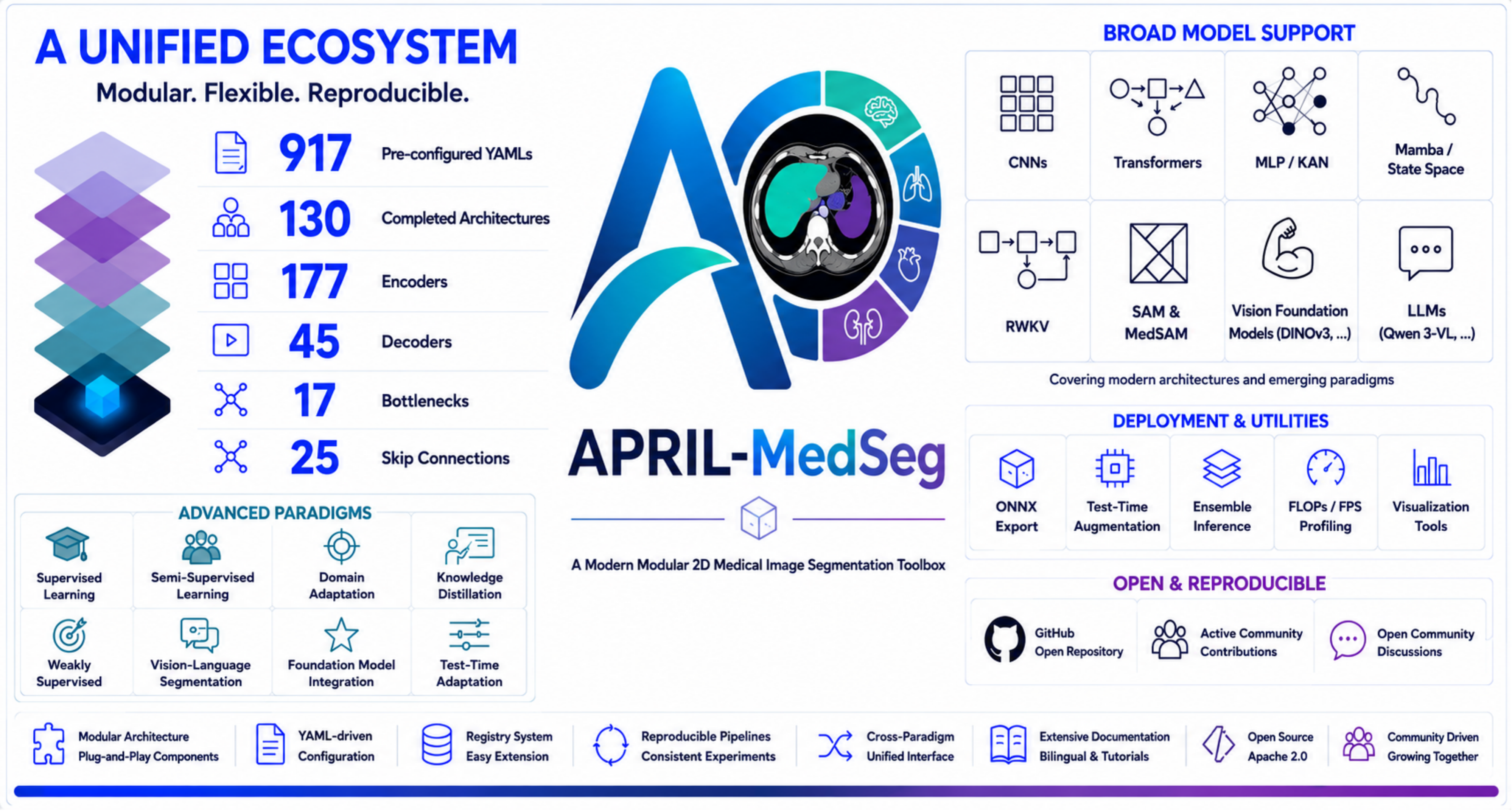}
    \caption{
    \textbf{APRIL-MedSeg}, a unified modular framework for medical image segmentation that integrates architectures, training paradigms, and deployment pipelines under a configuration-driven system.
    }
    \label{fig:teaser}
\end{figure*}

\section{Related Work}
\subsection{Segmentation Frameworks and Toolbox.} PaddleSeg~\cite{liu2021paddleseg} is a comprehensive deep learning segmentation toolkit built on PaddlePaddle. It provides an end-to-end framework for model development, training, and deployment across various vision tasks. The project is stably maintained, with its last major feature updates completed in late 2024. MMSegmentation~\cite{mmseg2020} is a highly modular, PyTorch-based segmentation framework, which features a config-driven paradigm and a large model zoo. Its core architecture stabilized after 2023. MONAI~\cite{cardoso2022monai} is a framework for healthcare imaging. It provides standardized segmentation networks alongside domain-specific preprocessing and evaluation tools. It remains a highly active project with continuous major updates into 2026. MIST~\cite{celaya2024mist}is a scalable, end-to-end framework for 3D medical image segmentation. The project is actively developed, with significant updates continuing into 2026. PaddleSeg-MedicalSeg~\cite{liu2021paddleseg} is a PaddleSeg extension specifically for 3D volumetric segmentation. The module matured following its late-2022 release. Segmentation Models PyTorch (SMP)~\cite{Iakubovskii:2019} is a popular, high-level PyTorch library primarily focused on 2D image segmentation, combining classic segmentation architectures with a vast collection of pre-trained encoders. However, no existing framework unifies modular architecture design, modern training paradigms, and configuration-driven reproducibility in a single system.
\subsection{Advances in Medical Image Segmentation}

The landscape of medical image segmentation has undergone a profound transformation in recent years, driven by rapid architectural innovations, the paradigm shift toward foundation models, and the continuous evolution of data-efficient training strategies.

\paragraph{Architectural Evolution: From CNNs to Linear-Complexity Sequence Models.} 
Historically, Convolutional Neural Networks (CNNs), epitomized by the U-Net family~\cite{ronneberger2015u}, established the baseline for dense prediction tasks. Subsequently, Vision Transformer (ViTs)~\cite{dosovitskiy2020image} introduce global receptive fields via self-attention mechanisms, significantly improving the delineation of complex anatomical structures~\cite{chen2021transunet, cao2022swin, heidari2023hiformer,liu2025cswin}.  Recently, the field has witnessed a surge in state-space models (SSMs) like Mamba~\cite{gu2023mamba} and linear RNN architectures such as RWKV~\cite{peng2023rwkv}. These next-generation architectures elegantly map sequence modeling into continuous-time dynamic systems or linear attention variants, achieving a global receptive field with linear computational complexity. Their integration into medical vision (e.g.,VM-UNet~\cite{ruan2024vm}, U-Mamba~\cite{ma2024u}, RWKV-UNet~\cite{jiang2025rwkv}) has redefined the efficiency-accuracy tradeoff, enabling the processing of extensive spatial contexts without prohibitive memory overhead.  Other efficient global modeling paradigms are also reshaping segmentation architectures. Linear-attention and recurrent-style frameworks, such as TTT-UNet~\cite{zhou2024ttt} and xLSTM-UNet~\cite{chen2024xlstm}, further reduce the quadratic complexity of conventional self-attention while preserving long-range dependency modeling. Meanwhile, MLP- and KAN-based architectures, including UNext~\cite{valanarasu2022unext}, Rolling-UNet~\cite{liu2024rolling}, and UKAN~\cite{moradzadeh2024ukan}, explore alternative token-mixing operators and nonlinear function approximators beyond standard convolution and attention mechanisms, reflecting a broader trend toward lightweight yet globally aware segmentation models.

\paragraph{The Rise of Foundation Models in General and Medical Vision.} 
The advent of large-scale pre-training has fundamentally disrupted traditional from-scratch learning paradigms. Generic vision and multi-modal foundation models, including the Segment Anything Model (SAM)~\cite{kirillov2023segment}, DINO series~\cite{caron2021emerging, li2025meddinov3, simeoni2025dinov3} and CLIP ~\cite{radford2021learning}, have demonstrated unprecedented zero-shot and few-shot generalization capabilities. Recognizing the domain gap between natural and clinical images, the community has rapidly developed medical-specific foundation models like MedSAM~\cite{ma2024segment} and BiomedCLIP~\cite{zhang2023biomedclip}. These models provide highly discriminative feature spaces, serving as powerful encoders that dramatically accelerate the convergence and boost the performance of downstream segmentation tasks. Furthermore, the recent proliferation of general and medical Multimodal Large Language Models (MLLMs)~\cite{bai2025qwen3vltechnicalreport, wang2025internvl3_5, lasateam2025lingshugeneralistfoundationmodel, jiang2025hulu} has revealed that their pre-trained vision modules inherently function as exceptionally robust visual encoders. Collectively, these models provide highly discriminative feature spaces, serving as powerful encoders that dramatically accelerate the convergence and boost the performance of downstream segmentation tasks.

\paragraph{Advanced Paradigms.}
Acquiring pixel-perfect annotations from clinical experts remains notoriously expensive and time-consuming, motivating the rapid development of advanced training paradigms for medical image segmentation. Semi-supervised learning methods, such as Mean Teacher~\cite{tarvainen2017mean}, FixMatch~\cite{sohn2020fixmatch}, and UniMatch~\cite{yang2023revisiting}, exploit large volumes of unlabeled data through consistency regularization and pseudo-labeling strategies. Weakly supervised segmentation further reduces annotation costs by leveraging coarse supervision signals, including bounding boxes~\cite{tian2021boxinst}, scribbles~\cite{lin2016scribblesup}, image-level labels via CAM-based localization~\cite{zhou2016learning,wang2020self}, and interactive point prompts~\cite{bearman2016s}. Beyond annotation efficiency, improving model generalization across hospitals, scanners, and imaging modalities has become another major research direction. Domain adaptation techniques, including adversarial alignment~\cite{ganin2016domain,vu2018advent}, frequency-domain adaptation~\cite{yang2020fda}, and self-training strategies~\cite{zou2018unsupervised}, have demonstrated strong capability in mitigating domain shifts. Meanwhile, knowledge distillation methods~\cite{romero2015fitnetshintsdeepnets,zhao2022decoupled,park2019relational} are increasingly adopted to compress cumbersome foundation or ensemble models into lightweight deployment-friendly networks while preserving segmentation accuracy. More recently, the field has expanded beyond conventional mask supervision toward text-guided segmentation paradigms. Vision-language models such as CRIS~\cite{wang2022cris}, LViT~\cite{li2023lvit}, and BiomedParse~\cite{zhao2024biomedparse} align radiological language with visual representations to enable semantically guided segmentation. In parallel, emerging zero-shot ``Detect-then-Segment'' pipelines combine MLLM-based grounding with promptable SAM-style segmenters, enabling open-vocabulary segmentation of unseen anatomical structures and pathologies without task-specific training data.

\begin{table*}[!t]
\centering
\scriptsize
\renewcommand{\arraystretch}{1.2}

\caption{Comparison of APRIL-MedSeg with Existing Segmentation Frameworks}
\label{tab:comparison}

\begin{tabularx}{\textwidth}{@{}lXXXXX@{}}
\toprule
Framework
& Medical Focus
& Model Zoo
& Loss Functions
& Decoupled Module\\
\midrule

Segmentation Models PyTorch~\cite{Iakubovskii:2019}
& \xmark & 12 & 8 & \cmark \\

PaddleSeg~\cite{liu2021paddleseg}
& \xmark & \textasciitilde66 & 22 & \cmark\\

PaddleSeg-MedicalSeg~\cite{liu2021paddleseg}
& \cmark & 6 & 3 & \xmark \\

MMSegmentation~\cite{mmseg2020}
& \xmark & \textasciitilde40 & 10 & \cmark \\

MONAI~\cite{cardoso2022monai}
& \cmark & \textasciitilde22 & \textasciitilde22 & \xmark \\

MIST~\cite{celaya2024mist}
& \cmark & 6 & 8 & \xmark\\

APRIL-MedSeg (ours)
& \cmark & 130 & 81 & \cmark \\
\bottomrule
\end{tabularx}

\vspace{1em}

\begin{tabularx}{\textwidth}{@{}lXXXXX@{}}
\toprule
Framework
& Foundation Models
& Advanced Paradigms
& Text-Guided
& Dataset Support \\
\midrule

Segmentation Models PyTorch~\cite{Iakubovskii:2019}
& \xmark & \xmark & \xmark & 3 \\

PaddleSeg~\cite{liu2021paddleseg}
& \cmark & \xmark & \cmark & 14 \\

PaddleSeg-MedicalSeg~\cite{liu2021paddleseg}
& \xmark & \xmark & \xmark & 6 \\

MMSegmentation~\cite{mmseg2020}
& \cmark & \xmark & \cmark & 21 \\

MONAI~\cite{cardoso2022monai}
& \xmark & \xmark & \xmark & \textasciitilde14 \\

MIST~\cite{celaya2024mist}
& \xmark & \xmark & \xmark & / \\

APRIL-MedSeg (ours)
& \cmark & \cmark & \cmark & 25 \\
\bottomrule
\end{tabularx}

\end{table*}

\section{APRIL-MedSeg Framework}
As shown in Table~\ref{tab:comparison}, existing segmentation frameworks are specialized but remain fragmented across different aspects. It is also worth noting that several widely used frameworks have seen limited major updates beyond 2024, with model expansion and paradigm coverage gradually plateauing. In contrast, APRIL-MedSeg provides a unified and extensible ecosystem that integrates modern segmentation architectures, training paradigms, and deployment workflows within a single configurable framework, enabling reproducible research and systematic benchmarking. To minimize implementation discrepancies and improve reproducibility, the framework aims to remain as faithful as possible to original official implementations rather than heavily simplified re-implementations.
\subsection{Modular Architecture Design}

\subsubsection{Design Philosophy}
The core insight motivating APRIL-MedSeg is that the vast majority of 2D medical segmentation networks can be structurally decoupled into four distinct, functionally independent modules. By formalizing this abstraction, we model the prediction pipeline as a multi-scale feature routing process:
\begin{equation}
\hat{y} = \mathcal{D}\Big(\mathcal{B}(f_L), \mathcal{S}(f_{1 \dots L-1})\Big)
\end{equation}
where $\{f_1, f_2, \dots, f_L\} = \mathcal{E}(x)$ represents the hierarchical feature pyramid extracted by the encoder. Within this unified design space:

\begin{itemize}[nosep,leftmargin=*]
    \item \textbf{Encoder} $\mathcal{E}$ progressively downsamples the input image $x$, extracting a multi-scale feature hierarchy that transitions from low-level fine-grained spatial details ($f_1$) to high-level dense semantic representations ($f_L$).
    \item \textbf{Bottleneck} $\mathcal{B}$ operates exclusively on the deepest, lowest-resolution feature map ($f_L$) to capture global context and model long-range dependencies, serving as the semantic core of the network.
    \item \textbf{Skip Connection} $\mathcal{S}$ acts as a controllable routing mechanism, transferring and optionally refining intermediate encoder features ($f_{1 \dots L-1}$) to mitigate spatial information loss caused by downsampling.
    \item \textbf{Decoder} $\mathcal{D}$ progressively upsamples and fuses the global contextual features from the bottleneck with the preserved spatial features from the skip connections, ultimately reconstructing the dense pixel-level segmentation map $\hat{y}$.
\end{itemize}

This formulation provides a unified abstraction, demonstrating that highly diverse architectures, from classic CNNs to modern models, can be elegantly interpreted as different instantiations of these four components.

\subsubsection{Registry and Configuration System}
To enable fully composable and reproducible experiment management, APRIL-MedSeg adopts a unified registry--configuration paradigm. Every component, from architectural modules to training objectives and data transforms, is registered via a Python decorator (\texttt{@REGISTRY.register()}) and instantiated purely through YAML declarations, eliminating the need for task-specific training scripts.

The framework maintains six global registries spanning the entire segmentation pipeline. \emph{Architecture registries} govern the four decoupled modules (encoder, decoder, skip connection, bottleneck), with the encoder registry further supporting dynamic resolution of any timm model via a \texttt{timm\_} prefix convention, granting access to hundreds of additional backbones without explicit registration. The \emph{loss registry} encompasses pixel-level, boundary-aware, distribution-alignment, and compound losses, where a recursive composition mechanism allows arbitrarily nested loss combinations with per-component weighting and automatic deep supervision injection. The \emph{augmentation registry} covers geometric, pixel-level, masking, and sample-level transforms, each exposing stochastic intensity parameters for reproducible yet diverse pipelines. On top of these registries, a hierarchical configuration inheritance mechanism enables child configs to override only the fields that differ from a parent, dramatically reducing redundancy.

The following example illustrates a representative configuration that composes a ResNet-50 encoder with ASPP bottleneck, attention-guided decoder, and a compound CE+Dice loss with deep supervision---all declared declaratively without writing any training code.

\lstdefinelanguage{yaml}{
  keywords={true,false,null},
  keywordstyle=\color{blue}\bfseries,
  basicstyle=\ttfamily\small,
  sensitive=false,
  comment=[l]{\#},
  commentstyle=\color{yamlcomment},
  moredelim=[l][\color{yamlkey}]{:},
  morestring=[b]',
  morestring=[b]"
}

\begin{tcolorbox}[
  colback=yamlbg,
  colframe=black!15,
  arc=2mm,
  boxrule=0.3pt,
  left=2mm,right=2mm,top=1mm,bottom=1mm,
  title={Representative YAML Configuration},
  fonttitle=\bfseries
]

\begin{lstlisting}[language=yaml]
model:
  num_classes: 9
  img_size: 224

  encoder:
    name: timm_resnet50
    pretrained: true

  decoder:
    name: attention

  skip_connection:
    name: concat

  bottleneck:
    name: aspp

training:
  epochs: 200
  batch_size: 24

  loss:
    name: compound
    params:
      losses:
        - name: ce
          weight: 0.4
        - name: dice
          weight: 0.6
\end{lstlisting}

\end{tcolorbox}

Once a configuration is specified, the entire pipeline can be launched with a single command. 
The framework automatically handles model construction, data loading, distributed training, 
loss composition, checkpointing, and evaluation. 
Different paradigms share a unified command-line interface with minimal configuration changes, 
as illustrated below:

\begin{tcolorbox}[
colback=gray!5,
colframe=black!15,
arc=1.5mm,
boxrule=0.3pt,
left=1mm,right=1mm,top=0.5mm,bottom=0.5mm
]
\small
\texttt{python train.py --config transunet.yaml} \hfill Supervised

\texttt{python semi\_train.py --config mean\_teacher.yaml} \hfill Semi-supervised

\texttt{python train\_domain\_adaptation.py --config advent.yaml} \hfill Domain adaptation

\texttt{python train\_distillation.py --teacher teacher.yaml --student student.yaml} \hfill Distillation

\texttt{python test.py --config transunet.yaml --tta} \hfill Inference + TTA
\end{tcolorbox}

\subsection{Architectures Support}
\subsubsection{Independent Networks}
APRIL-MedSeg covers 130 representative segmentation architectures organized into eight major paradigm families, reflecting the evolution from convolution-based designs to modern sequence modeling and vision-language systems. These families are defined based on their underlying computational mechanisms and modeling assumptions: CNN-based (36), Transformer-based (36), Mamba/SSM-based (24), RWKV-based (5), Other linear attention-based (3), KAN/MLP-based (4), SAM-family (10), and text-guided segmentation models (13).

\begin{table}[htp]
\centering
\caption{Complete list of the 130 independent segmentation architectures in APRIL-MedSeg, organized into eight paradigm families.}
\label{tab:networks}
\resizebox{\textwidth}{!}{
\begin{tabular}{l p{0.80\textwidth}}
\toprule
\textbf{Family} & \textbf{Architectures} \\
\midrule
CNN (36) & UNet~\cite{ronneberger2015u}, Attention U-Net~\cite{oktay2018attention}, UNet++~\cite{zhou2018unet++}, R2U-Net~\cite{alom2018recurrent}, MultiResUNet~\cite{ibtehaz2020multiresunet}, ResUNet-a~\cite{diakogiannis2020resunet}, ResUNet++~\cite{kaur2024resunet++}, UNet 3+~\cite{huang2020unet}, DenseUNet~\cite{cao2020denseunet}, scSE-UNet~\cite{liu2024scse}, SA-UNet~\cite{guo2021sa}, KiU-Net~\cite{valanarasu2020kiu}, PAN, LinkNet~\cite{chaurasia2017linknet}, PSPNet~\cite{zhao2017pyramid}, FR-UNet~\cite{liu2022full}, DCSAU-Net~\cite{xu2023dcsau}, CFA-Net~\cite{zhou2023cross}, MedNeXt~\cite{roy2023mednext}, nnU-Net~\cite{isensee2021nnu}, ACC-UNet~\cite{ibtehaz2023acc}, CMUNeXt~\cite{tang2024cmunext}, MEW-UNet~\cite{ruan2022mew}, LV-UNet~\cite{jiang2024lv}, EGE-UNet~\cite{ruan2023ege}, MALUNet~\cite{ruan2022malunet}, Lite-UNet~\cite{li2024lite}, MK-UNet~\cite{rahman2025mk}, U-Lite~\cite{dinh20231m}, AAU-Net~\cite{chen2022aau}, CMU-Net~\cite{tang2023cmu}, DconnNet~\cite{yang2023directional}, Polyper~\cite{shao2024polyper}, HoverNet Lite~\cite{graham2019hover},  DoubleU-Net~\cite{jha2020doubleu}, DSCNet~\cite{qi2023dynamic}\\
\midrule
Transformer (36) & SegFormer~\cite{xie2021segformer}, TransUNet~\cite{chen2021transunet}, Swin-UNet~\cite{cao2022swin}, Medical Transformer~\cite{valanarasu2021medical}, DAEFormer~\cite{azad2023daeformer}, MISSFormer~\cite{huang2021missformer}, H2Former~\cite{he2023h2former}, HiFormer~\cite{heidari2023hiformer}, MCTrans~\cite{ji2021multi}, MT-UNet~\cite{wang2022mixed}, ScaleFormer~\cite{huang2022scaleformer}, FAT-Net~\cite{wu2022fat}, nnFormer~\cite{zhou2023nnformer}, TransFuse~\cite{zhang2021transfuse}, LeViT-UNet~\cite{xu2023levit}, TransAttUNet~\cite{chen2023transattunet}, DA-TransUNet~\cite{sun2024transunet}, DS-TransUNet~\cite{lin2022ds}, UCTransNet~\cite{wang2022uctransnet}, Mobile-UViT~\cite{tang2025mobile}, CSWin-UNet~\cite{liu2025cswin}, FCBFormer~\cite{sanderson2022fcn}, PVT-UNet~\cite{xiong2024pvt}, PVTv2-b2-EMCAD~\cite{rahman2024emcad}, TransNetR~\cite{jha2024transnetr}, Polyp-PVT~\cite{dong2021polyp}, PVTv2-b2CASCADE~\cite{rahman2023medical}, HSNet~\cite{zhang2022hsnet}, SSFormer~\cite{shi2022ssformer}, LDNet~\cite{zhang2022lesion}, ESFPNet~\cite{chang2023esfpnet}, MIST~\cite{rahman2024mist}, SEPNet~\cite{wang2024polyp}, CTNet~\cite{xiao2024ctnet}, NuLite~\cite{tommasino2026nulite}, TransNuSeg~\cite{he2023transnuseg} \\
\midrule
Mamba/SSM (24) & Mamba-UNet~\cite{wang2024mamba}, H-vmunet~\cite{wu2025h}, LightM-UNet~\cite{liao2024lightm}, Swin-UMamba~\cite{liu2024swin}, U-Mamba~\cite{ma2024u}, UltraLight VM-UNet~\cite{wu2025ultralight}, VM-UNet~\cite{ruan2024vm}, VM-UNet V2~\cite{zhang2024vm}, LKM-UNet~\cite{wang2024lkm}, LoG-VMamba~\cite{dang2024log}, VMKLA-UNet~\cite{su2025vmkla}, UltraLBM-UNet~\cite{fan2025ultralbm}, nnMamba~\cite{gong2025nnmamba}, Polyp-Mamba~\cite{xu2024polyp}, HC-Mamba~\cite{xu2024hc}, AC-MambaSeg~\cite{nguyen2024ac}, DCM-Net~\cite{liu2026dcm}, DermoMamba~\cite{hoang2025dermomamba}, MUCM-Net~\cite{yuan2024mucm}, Serp-Mamba~\cite{wang2025serp}, SkinMamba~\cite{zou2024skinmamba}, MambaVesselNet++~\cite{xu2025mambavesselnet++}, ViM-UNet~\cite{archit2024vim}, UU-Mamba~\cite{tsai2024uu} \\
\midrule
RWKV (5) & RWKV-UNet~\cite{jiang2025rwkv}, U-RWKV (MICCAI)~\cite{ye2025u}, U-RWKV (TIP)~\cite{cai2026u}, MD-RWKV-UNet~\cite{fang2026md}, RIR-Zigzag~\cite{chen2025zig} \\
\midrule
Other Linear Attention (3) & TTT-UNet~\cite{zhou2024ttt}, xLSTM-UNet~\cite{chen2024xlstm}, U-VixLSTM~\cite{dutta2024vision} \\
\midrule
KAN/MLP (4) & UNeXt~\cite{valanarasu2022unext}, Rolling-UNet~\cite{liu2024rolling}, U-KAN~\cite{moradzadeh2024ukan}, WA-UKAN~\cite{liu2026wa} \\
\midrule
SAM family (10) & SAM~\cite{kirillov2023segment}, MedSAM~\cite{ma2024segment}, SAM 2~\cite{ravi2025sam}, MobileSAM~\cite{zhang2023faster}, SAMUS~\cite{lin2024beyond}, SAM-Med2D~\cite{cheng2023sam}, Medical SAM Adapter~\cite{wu2025medical}, SAMed~\cite{zhang2023customized}, AutoSAM~\cite{shaharabany2023autosam}, Lite-MedSAM~\cite{ma2024segment} \\
\midrule
Text-guided (12) & CLIP-Driven Referring Image Segmentation (CRIS)~\cite{wang2022cris}, BiomedParse~\cite{zhao2024biomedparse}, Language meets Vision Transformer (LViT)~\cite{li2023lvit}, Text-Guided Attention Network (TGANet)~\cite{tomar2022tganet}, Language-Guided Medical Segmentation (LanGuideMedSeg)~\cite{zhong2023languide}, Text-Promptable Region Segmentation (TPRO)~\cite{zhang2023tpro}, CausalCLIPSeg~\cite{chen2024causalclipseg}, CLIP-Driven Universal Model (CLIP-Universal)~\cite{liu2023clipuniversal}, CXR-CLIPSeg~\cite{you2023cxr}, Text-Promptable Diabetic Retinopathy Segmentation (TP-DRSeg)~\cite{li2024tpdrseg}, MedCLIP-SAM~\cite{koleilat2024medclipsam}, SaLIP~\cite{aleem2024salip} \\
\midrule
\textbf{Total} & \textbf{130} (118 standard + 12 text-guided) \\
\bottomrule
\end{tabular}
}
\end{table}

\subsubsection{Encoder}
The encoder registry provides 177 entries spanning nine categories by computational mechanism and modality specialization. \emph{CNN encoders} (11) provide strong local inductive bias through hierarchical convolution, suitable for small-scale datasets. \emph{Transformer encoders} (17) introduce global self-attention for long-range dependency modeling, including hybrid designs and pure vision transformers. \emph{Mamba/SSM encoders} (10) offer linear-complexity sequence modeling via state-space dynamics, while \emph{RWKV encoders} (5) combine transformer-style parallel training with linear-time recurrent inference. \emph{Linear-attention encoders} (5) replace quadratic attention with efficient approximations. \emph{KAN/MLP encoders} (4) employ learnable activation functions or pure token mixing. These encoders are adapted from existing standalone segmentation networks and backbone architectures. Additionally, a dynamic timm wrapper (85 pre-registered + 1000+ via prefix) exposes any model from models in Pytorch Image Models~\cite{grattafiori2024llama3herdmodels} (\texttt{timm.list\_models()} )via a simple \texttt{timm\_} prefix, granting access to hundreds of additional backbones without explicit registration.

Beyond hand-crafted encoders, the framework provides a large set of foundation model encoders covering multiple medical imaging modalities, including general vision, medical vision-language models, pathology, radiology, ophthalmology, dermatology, ultrasound, and endoscopy. These encoders are integrated through a unified DPT head that extracts multi-block multi-scale features from intermediate ViT layers, producing a consistent feature pyramid compatible with downstream decoders.

The framework supports multiple fine-tuning strategies, including \emph{frozen encoder} (only the decoder is trainable while preserving pre-trained representations), \emph{partial fine-tuning} (only the last layers are unfrozen), \emph{layer-wise learning rate decay} (progressively smaller learning rates for shallower layers), and \emph{full fine-tuning}. These strategies enable flexible control over the trade-off between preserving pre-trained knowledge and adapting to downstream medical tasks, which is particularly important in data-scarce medical imaging scenarios. The detailed composition of the supported foundation model encoders across different modalities is summarized in Table~\ref{tab:foundation_encoders}.
\begin{table*}[t]
\centering
\small
\caption{A unified set of 39 foundation model encoders across medical imaging modalities. All encoders are integrated via a unified DPT head that extracts multi-scale features from intermediate ViT blocks, producing a consistent feature pyramid for downstream decoders.}
\renewcommand{\arraystretch}{1.2}
\begin{tabularx}{\textwidth}{lX}
\toprule
\textbf{Category / Modality} & \textbf{Foundation Model Encoders} \\
\midrule

General Vision &
DINO~\cite{caron2021emerging}, DINOv2~\cite{oquab2023dinov2}, DINOv3~\cite{simeoni2025dinov3}, CLIP-ViT~\cite{radford2021learning}, SAM-ViT~\cite{ma2024segment} \\

General Medical &
BiomedCLIP~\cite{zhang2023biomedclip}, MedCLIP~\cite{wang2022medclip}, MedSigLIP~\cite{sellergren2025medgemma} \\

MLLM Vision Encoders &
Qwen2.5-VL~\cite{Qwen2.5-VL}, Qwen3-VL~\cite{bai2025qwen3vltechnicalreport}, MedGemma~\cite{sellergren2025medgemma}, LLaVA-Med~\cite{li2023llava}, HuatuoGPT-Vision~\cite{chen2024towards}, Lingshu~\cite{lasateam2025lingshugeneralistfoundationmodel}, HealthGPT~\cite{lin2025healthgpt}, Hulu-Med~\cite{jiang2025hulu} \\

Pathology &
Phikon~\cite{filiot2023scaling}, UNI~\cite{chen2024towardsuni}, PLIP~\cite{huang2023visual}, PhikonV2~\cite{filiot2024phikon}, MUSK~\cite{xiang2025vision}, KEEP~\cite{ZHOU2026777} \\

Radiology &
CheXzero~\cite{tiu2022expert}, BioViL~\cite{bannur2023learning}, Rad-DINO~\cite{perez2025exploring}, OmniRad~\cite{zedda2026omnirad} \\

Ophthalmology &
RETFound-MAE~\cite{zhou2023foundation}, 
RETFound-DINOv2~\cite{zhou2023foundation}, FLAIR~\cite{shi2025multimodal}, OphMAE~\cite{chang2026ophmae} \\

Dermatology &
MoNET~\cite{kim2024transparent}, DermCLIP~\cite{yan2025derm1m}, PanDerm~\cite{yan2025multimodal} \\

Ultrasound & UltraFedFM~\cite{jiang2025pretraining},
USF-MAE~\cite{MEGAHED2026110313}, SAMUS~\cite{lin2024beyond} \\

Endoscopy &
Endo-FM~\cite{wang2023foundation}, EndoViT~\cite{batic2024endovit}, SurgicalSAM~\cite{yue2024surgicalsam}\\

\bottomrule
\end{tabularx}
\label{tab:foundation_encoders}
\end{table*}
\subsubsection{Bottleneck}
The bottleneck registry provides 17 plug-in modules organized into seven categories according to their contextual modeling and feature refinement strategies. \emph{Baseline} bottlenecks provide either direct identity mapping or lightweight convolutional transformation. \emph{Multi-scale context} bottlenecks aggregate hierarchical receptive fields through pyramid pooling and dilated convolution operations. \emph{Channel attention} bottlenecks recalibrate feature importance along channel dimensions to enhance discriminative responses. \emph{Spatial-channel attention} bottlenecks jointly model spatial and channel-wise dependencies for more adaptive feature refinement. \emph{Self-attention / transformer \& hybrid} bottlenecks capture long-range dependencies through attention mechanisms or combine attention with convolutional operations. \emph{Position-aware convolution \& sparse expert} bottlenecks incorporate coordinate-aware representations or dynamic expert routing for adaptive feature processing. Finally, \emph{LLM-enhanced} bottlenecks utilize frozen pretrained large-model representations to enrich semantic understanding in segmentation pipelines.
\begin{table}[htp]
\centering
\caption{The 17 plug-in bottleneck modules in APRIL-MedSeg, organized into six functional categories.}
\label{tab:bottlenecks}
\resizebox{\textwidth}{!}{
\begin{tabular}{l p{0.80\textwidth}}
\toprule
\textbf{Category} & \textbf{Modules} \\
\midrule
Baseline (2) & Identity Pass-through (None), Basic Convolutional Block (Basic) \\
\midrule
Multi-scale Context (3) & Atrous Spatial Pyramid Pooling (ASPP)~\cite{chen2017deeplab}, DenseASPP~\cite{yang2018denseaspp}, Pyramid Pooling Module (PPM)~\cite{zhao2017pyramid} \\
\midrule
Channel Attention (2) & Squeeze-and-Excitation (SE)~\cite{hu2018squeeze}, Efficient Channel Attention (ECA)~\cite{wang2020eca} \\
\midrule
Spatial-Channel Attention (3) & Convolutional Block Attention Module (CBAM)~\cite{woo2018cbam}, Coordinate Attention (CA)~\cite{hou2021coordinate}, Spatial-Channel Attention (SCA)~\cite{chen2017sca} \\
\midrule
Self-Attention / Transformer \& Hybrid (4) & Self-Attention~\cite{vaswani2017attention}, Dual Attention~\cite{fu2019dual}, Gated Self-Attention, ACmix~\cite{pan2022integration} \\
\midrule
Position-aware Conv \& Sparse Expert (2) & Coordinate Convolution~\cite{liu2018coordconv}, Mixture-of-Experts (MoE)~\cite{shazeer2017outrageously} \\
\midrule
LLM-enhanced (1) & LLM4Seg~\cite{tang2025pre}  \\
\midrule
\textbf{Total} & \textbf{17} \\
\bottomrule
\end{tabular}
}
\end{table}
\subsubsection{Decoder}

The decoder registry provides 45 implementations organized into nine categories according to their feature fusion and upsampling strategies. \emph{Basic upsampling} decoders perform direct resolution recovery through interpolation or transposed convolution. \emph{Dense connection} decoders enhance multi-scale information flow via densely connected skip pathways. \emph{Cascade refinement} decoders progressively refine segmentation features across multiple stages. \emph{Pyramid aggregation} decoders integrate hierarchical contextual representations for multi-scale understanding. \emph{MLP-based} decoders employ lightweight perceptron-style feature mixing instead of convolution-heavy designs. \emph{Transformer-based} decoders utilize self-attention mechanisms for long-range dependency modeling and feature interaction. \emph{Attention-based} decoders selectively emphasize informative spatial or channel responses during feature fusion. \emph{Mamba-based} decoders introduce state-space modeling for efficient long-sequence representation learning. Finally, \emph{network-specific} decoders are customized for particular architectures and specialized design paradigms.
\begin{table}[htp]
\centering
\caption{The 45 decoder implementations  in APRIL-MedSeg, organized into nine functional categories.}
\label{tab:decoders}
\resizebox{\textwidth}{!}{
\begin{tabular}{l p{0.80\textwidth}}
\toprule
\textbf{Category} & \textbf{Decoders} \\
\midrule
Basic Upsampling (4) & Bilinear Upsampling, ConvTranspose Up-Cat, ConvTranspose Cat-Up, Depthwise-Separable \\
\midrule
Dense Connection (2) & UNet++~\cite{zhou2018unet++}, UNet 3+~\cite{huang2020unet} \\
\midrule
Cascade Refinement (10) & CASCADE~\cite{rahman2023medical}, CASCADE-Full~\cite{rahman2023medical}, CASCADE-EMCAD~\cite{rahman2023medical, rahman2024emcad}, CFM~\cite{dong2021polyp}, EMCAD~\cite{rahman2024emcad}, EDLDNet~\cite{dong2021polyp}, G-CASCADE~\cite{rahman2024g}, G-CASCADE-Cat~\cite{rahman2024g}, MERiT-Add~\cite{rahman2024multi}, MERiT-Cat~\cite{rahman2024multi} \\
\midrule
Pyramid Aggregation (2) & UPerNet~\cite{xiao2018unified}, DeepLabV3~\cite{chen2017rethinking} \\
\midrule
MLP-based (2) & MLP, SegFormer~\cite{xie2021segformer} \\
\midrule
Transformer-based (5) & DAEFormer~\cite{azad2023daeformer}, MISSFormer~\cite{huang2021missformer}, MT-UNet~\cite{wang2022mixed}, nnFormer~\cite{zhou2023nnformer}, Swin-UNet~\cite{cao2022swin}\\
\midrule
Attention-based (6) & Attention U-Net~\cite{oktay2018attention}, BA-Net~\cite{chen2019boundary}, CCNet~\cite{huang2019ccnet,huang2020ccnet}, Lawin~\cite{yan2022lawin}, OCRNet~\cite{yuan2020object}, UCTransNet~\cite{wang2022uctransnet} \\
\midrule
Mamba-based (1) & VM-UNet~\cite{ruan2024vm} \\
\midrule
Network-specific (13) &  CFA-Net~\cite{zhou2023cross}, DCSAU-Net~\cite{xu2023dcsau}, EGE-UNet~\cite{ruan2023ege}, FAT-Net~\cite{wu2022fat}, FFParser~\cite{gu2025transdiffseg}, H2Former~\cite{he2023h2former}, HAM~\cite{guo2022segnext}, HiFormer~\cite{heidari2023hiformer}, KiU-Net~\cite{valanarasu2020kiu}, MALUNet~\cite{ruan2022malunet}, RWKV-UNet~\cite{jiang2025rwkv}, ScaleFormer~\cite{huang2022scaleformer}, TransUNet~\cite{chen2021transunet}\\
\midrule
\textbf{Total} & \textbf{45} \\
\bottomrule
\end{tabular}
}
\end{table}
\subsubsection{Skip Connections}
The skip connection registry provides 25 implementations organized into five categories according to their feature transmission and fusion mechanisms. \emph{Basic} skip connections directly propagate encoder features through concatenation, addition, or dense routing. \emph{Attention-based} skip connections employ spatial, channel, or hybrid gating mechanisms to selectively emphasize informative representations during encoder--decoder fusion. \emph{Transformer-based} skip connections leverage cross-attention and token interaction mechanisms for adaptive multi-scale feature aggregation. \emph{Mamba-based} skip connections introduce state-space modeling for efficient long-range dependency propagation across hierarchical features. Finally, \emph{CNN fusion} skip connections utilize convolutional refinement and multi-scale alignment modules to enhance local structural consistency and feature integration.

\begin{table}[htp]
\centering
\caption{The 25 skip connection implementations in APRIL-MedSeg, organized into five functional categories.}
\label{tab:skip_connections}
\resizebox{\textwidth}{!}{
\begin{tabular}{l p{0.80\textwidth}}
\toprule
\textbf{Category} & \textbf{Skip Connections} \\
\midrule
Basic (3) & Concatenation, Element-wise Addition, Dense Skip~\cite{zhou2018unet++} \\
\midrule
Attention (10) & Attention Gate~\cite{oktay2018attention}, Channel Attention Block (CAB)~\cite{ruan2022malunet}, Spatial Attention Block (SAB)~\cite{ruan2022malunet}, Spatial-Channel SE~\cite{roy2018concurrent}, CBAM~\cite{woo2018cbam}, Gating, GRU Gate~\cite{cho2014learning}, Group Aggregation Bridge (GAB)~\cite{ruan2023ege}, SC-Att Bridge~\cite{ruan2022malunet}, Task-Adaptive Mixture of Skip Connections~\cite{luo2025rethinking}\\
\midrule
Transformer (5) & Cross Attention~\cite{vaswani2017attention}, Transformer Fusion~\cite{vaswani2017attention}, Aggregation Attention~\cite{rahman2023medical}, MISSFormer~\cite{huang2021missformer}, UCTrans skip~\cite{wang2022uctransnet}\\
\midrule
Mamba (1) & SK-VM++~\cite{wu2025sk} \\
\midrule
CNN Fusion (6) & BiFusion~\cite{zhang2021transfuse}, Deformable Convolution~\cite{dai2017deformable}, MultiScale Fusion, Feature Refine, Cross Channel Module, Scale-Diverse Integration(SDI)~\cite{peng2025u}  \\
\midrule
\textbf{Total} & \textbf{25} \\
\bottomrule
\end{tabular}
}
\end{table}

\subsubsection{Pretrain Weights}
To eliminate the friction of manual weight management, APRIL-MedSeg provides a four-tier automatic download system: (1)~a \texttt{WEIGHT\_REGISTRY} that hosts model-specific weights (SwinUNet~\cite{cao2022swin}, TransUNet~\cite{chen2021transunet}, H2Former~\cite{he2023h2former}, Mamba-UNet~\cite{wang2024mamba}, RWKV-UNet~\cite{jiang2025rwkv}, \etc) with multi-source fallback across GitHub, Google Cloud Storage, and HuggingFace; (2)~runtime integration with timm and torchvision that transparently downloads backbone weights when \texttt{pretrained: true} is set; (3)~dedicated SAM-family auto-download for ViT and SAM checkpoints; and (4)~HuggingFace Hub integration for 39 foundation model encoders, which auto-download via \texttt{transformers} and \texttt{open\_clip} at runtime. When all remote sources fail, the framework provides clear error messages listing the manual download URL, the exact cache path, and the YAML override key (\texttt{pretrained\_path}), ensuring that no experiment is blocked by network issues.

\subsection{Advanced Paradigms}

Beyond architecture design, APRIL-MedSeg organizes training strategies into a unified paradigm system that addresses data scarcity, domain shift, and supervision diversity in medical image segmentation, together with a comprehensive library of 81 loss functions.

These paradigms are grouped into three categories:

\begin{itemize}[nosep,leftmargin=*]

\item \textbf{Data-efficient learning paradigms:} including semi-supervised learning (20 methods) and weakly supervised segmentation (20 methods), which reduce reliance on dense annotations by leveraging unlabeled data or weak labels. Semi-supervised methods span consistency regularization, pseudo-labeling, dual-network co-training, and hybrid approaches. Weakly supervised methods support four annotation types, known as bounding boxes, scribbles, image-level labels, and point clicks, enabling segmentation with varying levels of annotation granularity. Representative methods are summarized in Table~\ref{tab:paradigms_cont}.

\item \textbf{Generalization and compression paradigms:} including domain adaptation (18 methods) and knowledge distillation (27 methods). Domain adaptation aligns feature distributions across clinical centers and imaging modalities via adversarial alignment, frequency-domain matching, self-training with progressive selection, and prototype-based alignment. Knowledge distillation compresses cumbersome models into lightweight deployable networks through feature-level, logit-level, and relation-level supervision, along with medical-specific distillation strategies. Representative methods are detailed in Table~\ref{tab:paradigms}.

\item \textbf{Task expansion paradigms:} including text-guided segmentation (12 trainable models + inference pipeline), which enables segmentation driven by natural language descriptions rather than pixel-level masks alone. This paradigm encompasses two distinct approaches: (1) \emph{trainable vision-language models} that learn cross-modal alignment between radiological text and visual features through joint training of text and image encoders; and (2) a \emph{zero-shot Detect-then-Segment pipeline} that requires no task-specific training, instead chaining an MLLM-based grounding detector with a prompt-driven SAM-based segmenter for open-vocabulary inference. The two approaches are complementary: trainable models achieve higher accuracy on established benchmarks, while the pipeline offers immediate zero-shot generalization to novel anatomical structures and pathologies without any labeled data. Representative methods are detailed in Table~\ref{tab:paradigms_cont}.

\end{itemize}

\begin{table*}[htp]
\centering
\caption{Summary of Learning Paradigms and Representative Methods (Part 1).}
\label{tab:paradigms}
\resizebox{\textwidth}{!}{
\begin{tabular}{ll p{0.55\textwidth}}
\toprule
\textbf{Paradigm} & \textbf{Category / Strategy} & \textbf{Representative Methods} \\
\midrule
\multirow{10}{*}{\textbf{Data-efficient}}
& \multicolumn{2}{l}{\textit{Semi-supervised Learning (20 methods)}} \\
& \quad Consistency regularization & Mean Teacher~\cite{tarvainen2017mean}, Uncertainty-Aware Mean Teacher (UA-MT)~\cite{yu2019uncertainty}, UniMatch~\cite{yang2023revisiting}, Pi-Model~\cite{laine2016temporal}, Temporal Ensembling~\cite{laine2016temporal}, Interpolation Consistency Training (ICT)~\cite{verma2022interpolation}, Regularized Dropout (R-Drop)~\cite{liang2021rdropregularizeddropoutneural} \\
& \quad Pseudo-labeling & FixMatch~\cite{sohn2020fixmatch}, FlexMatch~\cite{zhang2021flexmatch}, FreeMatch~\cite{wang2022freematch}, SoftMatch~\cite{chen2023softmatch}, Pseudo-Label~\cite{lee2013pseudo} \\
& \quad Dual-network co-training & Cross-Consistency Training (CCT)~\cite{ouali2020semisupervisedsemanticsegmentationcrossconsistency}, Deep Co-Training~\cite{qiao2018deep}, Cross-Teaching~\cite{luo2022semi}, Cross Pseudo Supervision (CPS)~\cite{chen2021semisupervisedsemanticsegmentationcross} \\
& \quad Hybrid approaches & Uncertainty Rectified Pyramid Consistency (URPC)~\cite{luo2021efficientsemisupervisedgrosstarget}, AllSpark~\cite{wang2024allspark}, DiffRect~\cite{liu2024diffrect}, CorrMatch~\cite{sun2023corrmatchlabelpropagationcorrelation} \\
\cmidrule{2-3}
& \multicolumn{2}{l}{\textit{Weakly Supervised Segmentation (20 methods)}} \\
& \quad Bounding boxes & BoxSup~\cite{dai2015boxsup}, BoxInstance Segmentation (BoxInst)~\cite{tian2021boxinst} \\
& \quad Scribbles & ScribbleSup~\cite{lin2016scribblesup} \\
& \quad Image-level labels & Class Activation Mapping (CAM)~\cite{zhou2016learning}, Multi-Instance Learning (MIL)~\cite{ilse2018attention}, Tree-Structured Energy (TreeEnergy)~\cite{liang2022tree}, Self-supervised Equivariant Attention Mechanism (SEAM)~\cite{wang2020self}, Puzzle Class Activation Mapping (PuzzleCAM)~\cite{jo2021puzzle}, Adversarial Complementary CAM (AdvCAM)~\cite{lee2021anti}, Multi-Class Token Transformer (MCTformer)~\cite{xu2022multi}, Explicit Pseudo-label Supervision (EPS)~\cite{lee2021railroad}, Re-weighted CAM (ReCAM)~\cite{chen2022classreactivationmapsweaklysupervised}, Token Contrast (ToCo)~\cite{ru2023tokencontrastweaklysupervisedsemantic}, Low-Pass CAM (LPCAM)~\cite{chen2023extractingclassactivationmaps}, Model-Agnostic Biased Object Removal (MARS)~\cite{jo2023mars}, Dual Pseudo Label (DuPL)~\cite{wu2024dupldualstudenttrustworthy}, Momentum Refinement (MoRe)~\cite{yang2025moreclasspatchattention}, Pseudo-label Denoising with Prior Model (PSDPM)~\cite{zhao2024psdpm}, Semantic Pseudo Label Selection (SemPLeS)~\cite{lin2025semanticpromptlearningweaklysupervised} \\
& \quad Point clicks & PointSup~\cite{bearman2016s} \\
\bottomrule
\end{tabular}
}
\end{table*}

\begin{table*}[htp]
\centering
\caption{Summary of Learning Paradigms and Representative Methods (Continued).}
\addtocounter{table}{-1}
\label{tab:paradigms_cont}
\resizebox{\textwidth}{!}{
\begin{tabular}{ll p{0.55\textwidth}}
\toprule
\textbf{Paradigm} & \textbf{Category / Strategy} & \textbf{Representative Methods} \\
\midrule
\multirow{5}{*}{\textbf{Generalization}}
& \multicolumn{2}{l}{\textit{Domain Adaptation (18 methods)}} \\
& \quad Adversarial alignment & Domain-Adversarial Neural Network (DANN)~\cite{ganin2016domain}, Adversarial Entropy Minimization (AdvEnt)~\cite{vu2018advent} \\
& \quad Frequency-domain matching & Fourier Domain Adaptation (FDA)~\cite{yang2020fda}, Masked Image Consistency (MIC)~\cite{hoyer2023mic}, Dual-Domain Decoupled Bridging (DDB)~\cite{chen2022deliberateddomainbridgingdomain} \\
& \quad Self-training & Source Only, Test-Time Entropy Minimization (TENT)~\cite{wang2021tentfullytesttimeadaptation}, Dual Pseudo-Labeling (DPL)~\cite{chen2021sourcefreedomainadaptivefundus}, Class-Balanced Mean Teacher (CBMT)~\cite{tang2023sourcefreedomainadaptivefundus}, Confidence Regularized Self-Training (CRST)~\cite{zou2019confidence}, PixMatch~\cite{melaskyriazi2021pixmatchunsuperviseddomainadaptation}, DAFormer~\cite{hoyer2022daformerimprovingnetworkarchitectures}, High-Resolution Domain Adapter (HRDA)~\cite{hoyer2022hrdacontextawarehighresolutiondomainadaptive}, MICDrop~\cite{yang2024micdropmaskingimagedepth}, SemiVL~\cite{hoyer2023semivlsemisupervisedsemanticsegmentation} \\
& \quad Prototype-based alignment & Semantic-Guided Pixel Contrast (SePiCo)~\cite{xie2023sepico}, Distribution-Guided Alignment (DiGA)~\cite{shen2023diga}, Pixel- and Patch-wise Self-supervised Mixing (PiPa)~\cite{chen2023pipa} \\
\midrule
\multirow{5}{*}{\textbf{Compression}}
& \multicolumn{2}{l}{\textit{Knowledge Distillation (27 methods)}} \\
& \quad Feature-level supervision & FitNets~\cite{romero2015fitnetshintsdeepnets}, Attention Transfer (AT)~\cite{zagoruyko2016paying}, Flow of Solution Procedure (FSP)~\cite{yim2017gift}, Neuron Selectivity Transfer (NST)~\cite{huang2017likelikeknowledgedistill}, Variational Information Distillation (VID)~\cite{ahn2019variationalinformationdistillationknowledge}, Masked Generative Distillation (MGD)~\cite{yang2022masked}, Knowledge Review (ReviewKD)~\cite{chen2021distillingknowledgeknowledgereview}, Scale Decoupled Distillation (SDD)~\cite{luo2024scaledecoupleddistillation}, Attention Mimicry~\cite{zagoruyko2016paying}, UNet Distillation \\
& \quad Logit-level supervision & Vanilla Knowledge Distillation (Vanilla KD)~\cite{hinton2015distillingknowledgeneuralnetwork}, Decoupled Knowledge Distillation (DKD)~\cite{zhao2022decoupled}, Channel-Wise Distillation (CWD)~\cite{shu2021channel}, Distance-Wise Supervision (DIST)~\cite{huang2022knowledge}, Simple Knowledge Distillation (SimKD)~\cite{chen2022knowledgedistillationreusedteacher}, Normalized Knowledge Distillation (NORM)~\cite{chi2023normkdnormalizedlogitsknowledge}, Adaptive Inter-Class Similarity Distillation (AICSD)~\cite{mansourian2023aicsdadaptiveinterclasssimilarity}, Logit Standardization Knowledge Distillation (LSKD)~\cite{sun2024logitstandardizationknowledgedistillation}, Transformed Teacher Matching (TTM)~\cite{zheng2024knowledgedistillationbasedtransformed}, Curriculum Temperature Knowledge Distillation (CTKD)~\cite{li2022curriculumtemperatureknowledgedistillation}, Multi-Level Logit Distillation (MLKD)~\cite{jin2023multi} \\
& \quad Relation-level supervision & Relational Knowledge Distillation (RKD)~\cite{park2019relational}, Cross-Image Relational Knowledge Distillation (CIRKD)~\cite{yang2022cross} \\
& \quad Medical-specific & Anatomy Knowledge Distillation (Anatomy-KD), Boundary-aware Knowledge Distillation (Boundary-KD)~\cite{gong2021boundary}, Multi-Organ Knowledge Distillation (Multi-Organ-KD), Cross-Modality Knowledge Distillation (Cross-Modality-KD) \\
\midrule
\multirow{3}{*}{\textbf{Task Expansion}}
& \multicolumn{2}{l}{\textit{Text-guided Segmentation (12 models + pipeline)}} \\
& \quad Trainable VLM & CLIP-Driven Referring Image Segmentation (CRIS)~\cite{wang2022cris}, BiomedParse~\cite{zhao2024biomedparse}, Language meets Vision Transformer (LViT)~\cite{li2023lvit}, Text-Guided Attention Network (TGANet)~\cite{tomar2022tganet}, Language-Guided Medical Segmentation (LanGuideMedSeg)~\cite{zhong2023languide}, Text-Promptable Region Segmentation (TPRO)~\cite{zhang2023tpro}, CausalCLIPSeg~\cite{chen2024causalclipseg}, CLIP-Driven Universal Model (CLIP-Universal)~\cite{liu2023clipuniversal}, CXR-CLIPSeg~\cite{you2023cxr}, Text-Promptable Diabetic Retinopathy Segmentation (TP-DRSeg)~\cite{li2024tpdrseg}, MedCLIP-SAM~\cite{koleilat2024medclipsam}, SaLIP~\cite{aleem2024salip} \\
& \quad Zero-shot pipeline & Detect-then-Segment pipeline (Multimodal Large Language Model (MLLM) + Segment Anything Model (SAM)) \\
\bottomrule
\end{tabular}
}
\end{table*}

In addition, the framework supports a \textbf{unified loss composition system} with 81 loss functions and enables flexible combination of objectives with automatic deep supervision, allowing complex training objectives without modifying model code.
\subsection{Data and Augmentation}


\textbf{Dataset support.} 26 built-in medical datasets spanning 8 imaging
modalities: \emph{CT} (Synapse~\cite{landman2015synapse}, COVID-CT-Seg~\cite{ma2021covidct},MosMedData+~\cite{morozov2020mosmeddata}),
\emph{MRI} (ACDC~\cite{bernard2018acdc}), \emph{Chest X-ray}
(Montgomery~\cite{jaeger2014montgomery}, Shenzhen~\cite{jaeger2014montgomery},
QaTa-COV19~\cite{degerli2022osegnet}), \emph{Fundus photography}
(DRIVE~\cite{staal2004drive}, STARE~\cite{hoover2000stare},
CHASE\_DB1~\cite{fraz2012chase}, HRF~\cite{budai2013hrf}, ARIA~\cite{farnell2008aria},
RITE~\cite{hu2013rite}, REFUGE~\cite{orlando2020refuge},
Drishti-GS~\cite{sivaswamy2014drishti}), \emph{Dermoscopy}
(ISIC 2016~\cite{gutman2016skin}/2017~\cite{codella2018skin}/2018~\cite{codella2019skin},
PH2~\cite{mendoncca2013ph}), \emph{Endoscopy}
(CVC-ClinicDB~\cite{bernal2015wm}, CVC-ColonDB~\cite{tajbakhsh2015colondb},
Kvasir-SEG~\cite{jha2019kvasir}), \emph{Histopathology}
(GlaS~\cite{sirinukunwattana2017gland}, PanNuke~\cite{gamper2019pannuke},
MoNuSeg~\cite{kumar2017monuseg}), and \emph{Ultrasound} (BUSI~\cite{al2020dataset}).
Each dataset is accessible through 5 data loading types (binary, multi-class,
domain adaptation, semi-supervised, text-image pairs) with 4 split strategies:
explicit file paths, ratio-based random splits, $K$-fold cross-validation, and
predefined community splits.

\textbf{Augmentation pipeline.} 24 methods organized into 4 categories: geometric (flip, rotation, affine, perspective, elastic deform, scale), pixel-level (photometric distortion, color jitter, CLAHE, Gaussian blur/noise, solarize), masking (random erasing, coarse dropout~\cite{devries2017improved}, grid mask~\cite{chen2020gridmask}), and sample-level (copy-paste~\cite{ghiasi2021simple}, mosaic~\cite{bochkovskiy2020yolov4}). All intensity parameters use \texttt{\_range} suffixes for stochastic sampling within specified bounds, ensuring reproducible yet diverse augmentation via YAML configuration alone.

\subsection{Deployment and Efficiency}

\textbf{ONNX export}: Single-command conversion with dynamic spatial dimensions and ONNX Runtime verification for deployment-ready inference.

\textbf{Model profiling}: FLOPs, parameter count, and FPS measurement under configurable input resolutions, enabling systematic efficiency comparison across architecture combinations.

\textbf{Test-time augmentation (TTA)}: Supports multi-scale inference ($\{0.5\times, 0.75\times, 1.0\times, 1.25\times, 1.5\times\}$) combined with horizontal and vertical flipping, with configurable aggregation strategies (mean, max, or voting) to improve prediction robustness without retraining.

\textbf{Ensemble}: Multi-model prediction fusion that combines outputs from independently trained models via weighted averaging (per-model confidence weighting) or majority voting (per-pixel label consensus), both configurable through YAML without code changes. 

\section{Availability, Documentation, and Intended Users}
\label{sec:availability}

\paragraph{Software Availability.}
The code of APRIL-MedSeg is available at \url{https://github.com/juntaoJianggavin/APRIL-MedSeg} under an Apache 2.0 license. The repository includes complete source code, configuration templates, and comprehensive documentation for reproducible experiments. Dependencies are limited to widely used open-source packages, including PyTorch ($\geq$2.4), timm, MONAI, numpy, and opencv-python. Crucially, all integrated models and codebase have been thoroughly tested and will undergo continuous, iterative validation to guarantee long-term reliability and correctness.

\paragraph{Documentation.}
APRIL-MedSeg provides comprehensive bilingual (English/Chinese) documentation organized into five modules covering models, training paradigms, data, deployment, and research guidance. Each module includes API references, step-by-step usage guides, and YAML configuration examples. Beyond reference documentation, the framework offers a structured 9-chapter tutorial series that progressively guides readers from fundamentals to advanced topics: The tutorial system covers:

\begin{itemize}[leftmargin=*]

\item Introduction to medical image segmentation concepts, evaluation metrics, and method evolution;

\item U-Net architecture and its family variants;

\item Data formats, dataset split strategies, and augmentation pipelines;

\item Training workflows including loss composition, AMP/DDP acceleration, and evaluation protocols;

\item Encoder deep dive covering CNN, Transformer, Mamba, and RWKV backbones with timm integration;

\item Decoder taxonomy and skip connection design patterns;

\item Foundation model integration across nine medical imaging modalities with fine-tuning strategies;

\item Advanced training paradigms including semi-supervised learning, domain adaptation, knowledge distillation, and weakly supervised segmentation;

\item Deployment workflows including ONNX export, TTA, ensemble inference, and the MLLM inference pipeline.

\end{itemize}

The tutorials are designed as self-contained and beginner-friendly units, each combining conceptual explanations with runnable code snippets to bridge theory and practice.

\paragraph{Who it is for.} The framework serves multiple user communities: \emph{Architecture researchers} can rapidly prototype and benchmark new encoder, decoder, or bottleneck designs by registering a single Python class and composing it with existing modules via YAML. \emph{Paradigm researchers} working on semi-supervised, domain adaptation, or text-guided methods can leverage the built-in training pipelines and loss composition system without rewriting boilerplate code. \emph{Clinical practitioners} can select pre-configured YAML templates for their dataset and modality, obtaining reproducible baselines with a single command. \emph{Educators and students} benefit from the bilingual documentation and step-by-step tutorials that progressively introduce segmentation concepts from basic U-Net to foundation model integration.

\section{Future Plan}

We outline several directions for the continued development of APRIL-MedSeg. \textbf{Benchmarking and validation.} While large-scale benchmarking is not the primary focus as recent works have provided such comparison~\cite{tang2025u}, we will leverage the modular design to explore new architecture combinations and training settings, and provide representative empirical results to demonstrate the effectiveness of the framework and facilitate model selection. \textbf{Maintenance and updates.} We commit to active maintenance, including timely updates and issue resolution within a short iteration cycle. \textbf{Community growth.} We aim to foster an open-source community to encourage contributions of models, training paradigms, and deployment tools. \textbf{3D and video extension.} We plan to extend the framework to 3D and video medical image segmentation in the future, while acknowledging that this space is already well covered by mature frameworks such as nnU-Net~\cite{isensee2021nnu} and MONAI.
\section{Conclusion}
APRIL-MedSeg is an open-source framework for 2D medical image segmentation that unifies modern architectures, training paradigms, and evaluation workflows within a modular, YAML-driven system. By decoupling model design from engineering complexity, it enables reproducible research, fair benchmarking, and rapid prototyping under a consistent framework. Overall, it provides a general-purpose and extensible infrastructure to support research and development in the medical image segmentation community.

\bibliography{april_aigc}

\end{document}